\documentclass{article}

\usepackage[preprint]{neurips_2026}

\usepackage[utf8]{inputenc} %
\usepackage[T1]{fontenc}    %
\usepackage{hyperref}       %
\usepackage{url}            %
\usepackage{booktabs}       %
\usepackage{amsfonts}       %
\usepackage{nicefrac}       %
\usepackage{microtype}      %
\usepackage{xcolor}         %
\usepackage{graphicx}
\usepackage{amsmath}

\title{Efficiently Aligning Draft Models via Parameter- and Data-Efficient Adaptation}

\author{
\textbf{Luxi Lin\thanks{Equal contribution}} $^{1}$,
\textbf{Zhihang Lin\footnotemark[1]} $^{1,2}$,
\textbf{Zhanpeng Zeng\thanks{Corresponding author}} $^{1}$,
\textbf{Yuhao Chen} $^{3}$, \\
\textbf{Qingyu Zhang} $^{4}$,
\textbf{Jixiang Luo} $^{5}$,
\textbf{Rongrong Ji} $^{1}$ \\
$^1$Key Laboratory of Multimedia Trusted Perception and Efficient Computing, \\
Ministry of Education of China, Xiamen University, 361005, P.R. China \\
$^2$Shanghai Innovation Institute \\
$^3$University of Science and Technology of China (USTC) \\
$^4$University of Chinese Academy of Sciences (UCAS) \\
$^5$Institute of Artificial Intelligence (TeleAI), China Telecom \\
{\tt\small lewuluu@gmail.com, zzeng@xmu.edu.cn}
}

\begin{document}

\maketitle

\begin{abstract}
Speculative decoding accelerates LLM inference but suffers from performance degradation when target models are fine-tuned for specific domains.
A naive solution is to retrain draft models for every target model, which is costly and inefficient.
To address this, we introduce a practical parameter- and data-efficient framework named Efficient Draft Adaptation, abbreviated as EDA, for efficiently adapting draft models.
EDA introduces three innovations:
(1) a decoupled architecture that utilizes shared and private components to model the shared and target-specific output distributions separately, enabling parameter-efficient adaptation by updating only the lightweight private component;
(2) a data regeneration strategy that utilizes the fine-tuned target model to regenerate training data, thereby improving the alignment between training and speculative decoding, leading to higher average acceptance length;
(3) a sample selection mechanism that prioritizes high-value data for efficient adaptation.
Our experiments show that EDA effectively restores speculative performance on fine-tuned models, achieving superior average acceptance lengths with significantly reduced training costs compared to full retraining.
Code is available at \url{https://github.com/Lyn-Lucy/Efficient-Draft-Adaptation}.
\end{abstract}

\section{Introduction}
Large language models (LLMs) have achieved remarkable progress in natural language understanding and generation~\cite{radford2018gpt,touvron2023llama,bai2023qwen}.
However, their inference efficiency remains bottlenecked by the token-by-token nature of autoregressive decoding, limiting deployment in latency- and throughput-sensitive applications.
Speculative decoding~\cite{leviathan2023spd,chen2023spsampling} offers a lossless acceleration paradigm: a lightweight draft model proposes multiple candidate tokens ahead, and the target model verifies these candidates in parallel.
By adopting an accept–reject sampling procedure that guarantees the final output distribution is exactly preserved, speculative decoding reduces target-model forward passes without sacrificing generation quality, yielding substantial speedup in practice~\cite{kwon2023vllm,SGLang}.

\begin{figure}
  \centering
  \includegraphics[width=0.6\linewidth]{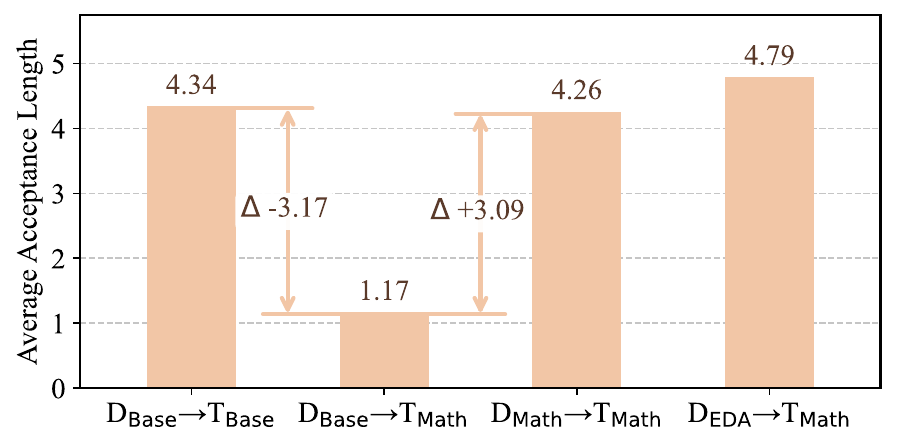}
  \caption{
  The comparison of average acceptance length on GSM8K under different draft-target pairings.
  We refer to the target models as $T_{Base}$ (Qwen2.5-7B) and $T_{Math}$ (Qwen2.5-Math-7B).
  The target model $T_{Base}$'s draft model $D_{Base}$ achieves high average acceptance length when paired ($D_{Base} \rightarrow T_{Base}$), while it suffers a substantial acceptance drop when paired with $T_{Math}$ ($D_{Base} \rightarrow T_{Math}$).
  Our EDA framework can restore performance through efficient, lightweight adaptation.
  }
\label{fig:transfer_gap}
\end{figure}

In practice, generic pre-trained LLMs are often further adapted to specific tasks or domains (e.g., math~\cite{yu2023metamath,yue2024mammoth}, code~\cite{roziere2023code,luo2023wizardcoder}, and medicine~\cite{wu2023pmedical,zhang2023huatuogpt}) via supervised fine-tuning (SFT), which substantially shifts their output distributions.
However, speculative decoding is effective only when the draft model’s output distribution closely aligns with that of the target model.
As a result, the draft model trained before SFT no longer aligns well with the fine-tuned target model, leading to degraded average acceptance length and speedup.
We illustrate this phenomenon in Figure~\ref{fig:transfer_gap}.
A draft model ($D_{Base}$) trained to match the pre-fine-tuning target model, denoted as $T_{Base}$, exhibits a substantial drop in average acceptance length when it is directly used as the draft model for the fine-tuned target model, denoted as $T_{Math}$.
Thus, we must retrain a dedicated draft model for each fine-tuned target model to realign their output distributions, incurring significant training costs.
There is thus a pressing need for efficient draft adaptation techniques that can quickly track target-model changes.

In this paper, we propose a novel approach to \textbf{E}fficiently align \textbf{D}raft models via parameter- and data-efficient \textbf{A}daptation, dubbed EDA.
Specifically, we observe that the output distributions of target models before and after fine-tuning share a substantial overlap, as evidenced by the non-zero average acceptance length when directly applying the draft model trained for the base target model to the fine-tuned target model, as shown in Figure~\ref{fig:transfer_gap}.
Thus, we decouple the draft model into shared and private components to separately model the shared and task-specific output distributions.
Compared to traditional methods that train a monolithic draft model for each target model, our design enables efficient transfer by reusing the shared component and updating only the lightweight private component when adapting to a new target model, showing the parameter efficiency of our approach.

Next, we improve the average acceptance length of the draft model by reducing the mismatch between the output distribution of training and speculative decoding stages.
Specifically, during the draft model training stage, the draft model is optimized to predict the next token in the provided dataset.
However, during speculative decoding, the draft model needs to predict the target model's next token, which may differ from the ground-truth token in the provided dataset, leading to a mismatch between training and speculative decoding.
To address this issue, we regenerate the training set using the fine-tuned target model itself, thereby improving the alignment between training and drafting, leading to a higher average acceptance length.

Finally, we further improve the adaptation efficiency from the data perspective, as adapting the draft model with a large training set can still incur unnecessary cost.
We found that not all training samples contribute equally to improving the average acceptance length during adaptation.
Thus, we introduce a sample value selection mechanism that scores and ranks candidate training samples to construct a more compact adaptation subset.
Under a fixed training budget, this enables the draft model to focus on generation segments that are most influential for improving average acceptance length during adaptation, thereby achieving higher adaptation efficiency with fewer samples.

We conduct extensive experiments on various downstream tasks to evaluate the effectiveness of EDA. For example, when adapting a draft model trained for Qwen2.5-7B to the fine-tuned target model Qwen2.5-Math-7B, EDA achieves an average acceptance length of $4.79$, significantly outperforming the baseline adaptation method that achieves only $4.37$, while incurring only $39.2\%$ of the training cost of full draft model retraining.

\section{Related Work}
\subsection{Speculative Decoding}
Speculative decoding~\cite{leviathan2023spd,chen2023spsampling,lin2025MSD} utilizes a draft model to generate multiple candidate tokens, which are then validated by a target model, significantly speeding up LLM decoding.
Its realized acceleration is largely governed by how well the draft model matches the target model’s conditional distribution, commonly reflected by the average acceptance length.
Prior work commonly improves speculative decoding by training or aligning a dedicated draft model for a particular target model, e.g., via distillation-style distribution matching in token space~\cite{cai2024medusa,zhou2023distillspec,yan2025dsd} or by drafting in an intermediate representation space as in EAGLE~\cite{li2024eagle,li2024eagle2}.
These methods typically optimize the draft model to better follow the target model’s next-token preferences, thereby increasing consecutive acceptances and translating into higher end-to-end speedup.

However, they generally assume a fixed target model and require re-training or re-adaptation when the target changes to preserve average acceptance length and speedup.
In contrast, we treat cross-target transfer of the draft model as a key problem, enabling reuse and efficient adaptation via a shared private decomposition, rather than training a separate draft model for each target model.

\subsection{Parameter-Efficient Fine-Tuning}
Parameter-efficient fine-tuning (PEFT) adapts large pre-trained models by updating only a small set of additional parameters while keeping the backbone frozen. Representative approaches include inserting lightweight adapters into Transformer blocks~\cite{houlsby2019parameter}, learning low-rank updates as in LoRA~\cite{hu2022lora}, and optimizing prompt-like parameters such as prefix/prompt tuning~\cite{li2021prefix}. More recently, PEFT has also been extended toward more modular designs that introduce multiple lightweight modules and allocate capacity more flexibly. For instance, \citet{yu2024boosting} proposes mixture-of-experts adapters with lightweight routing to support specialization and reduce interference in multi-task and continual learning.

Our work is related to parameter-efficient adaptation, which updates only a small set of parameters while keeping most of the model fixed~\cite{houlsby2019parameter,hu2022lora,li2021prefix}.
Unlike traditional PEFT methods that adapt the target model, we apply this approach to the speculative draft model, enabling the reuse of shared structures while updating only a small, target-specific component. This ensures the draft model efficiently aligns with changes in the target model's generation patterns, preserving acceptance rates and decoding speedup.

\section{Speculative Decoding}
\textbf{Draft Model Training Stage.} Starting with a prefix $x_{<t}$ sampled from the data distribution $\mathcal{D}$, the draft model $\theta_{d}$ is trained to closely match the target distribution of the larger model $\theta_{t}$ as:
\begin{equation}
\begin{split}
  \theta_{d} =  \arg\min_{\theta_d} \, & \mathbb{E}_{x \sim \mathcal{D}} \sum_{t} \Bigl[
    L_{reg} + \text{CE}\bigl(P_{\theta_{t}}(x_t \mid x_{<t}), P_{\theta_{d}}(x_t \mid x_{<t})\bigr)
  \Bigr],
\end{split}\label{eq:draft_training_loss}
\end{equation}
where CE denotes the cross-entropy loss, and $L_{reg}$ is a regularization term depending on specific implementations, such as intermediate representation matching~\cite{li2024eagle}.
$P_{\theta_{t}}(x_t \mid x_{<t})$ and $P_{\theta_{d}}(x_t \mid x_{<t})$ denote the target and draft model's prediction probability distributions, respectively.
When $P_{\theta_{t}}(x_t \mid x_{<t})$ shifts, as often happens in practice (e.g., after fine-tuning), the draft model $P_{\theta_{d}}(x_t \mid x_{<t})$ no longer aligns well with it.
This misalignment necessitates retraining the entire draft model on updated datasets for each fine-tuned target model, resulting in significant training overhead.

\textbf{Speculation Stage.} Based on the prefix $x_{<t}$, the draft model proposes  $K$ speculative candidate tokens $\hat{x}_{t:t+K-1}=\{x_t,\cdots, x_{t+K-1}\}$ from draft distributions.

\textbf{Verification Stage.}
The target model verifies the conditional distributions in parallel,
$P_{\theta_{t}}(x_{t+i} \mid x_{<t}, \hat{x}_{t:t+i-1})$ for $i=0,\ldots,K-1$,
and accepts candidate tokens based on the acceptance rule~\cite{leviathan2023spd} until encountering the first rejection.
At the rejection point, a replacement token is sampled directly from the target model's distribution. This mechanism ensures the final output distribution remains identical to standard autoregressive sampling from the target model.

\section{Method}
\begin{figure*}[t]
  \centering
  \includegraphics[width=\textwidth]{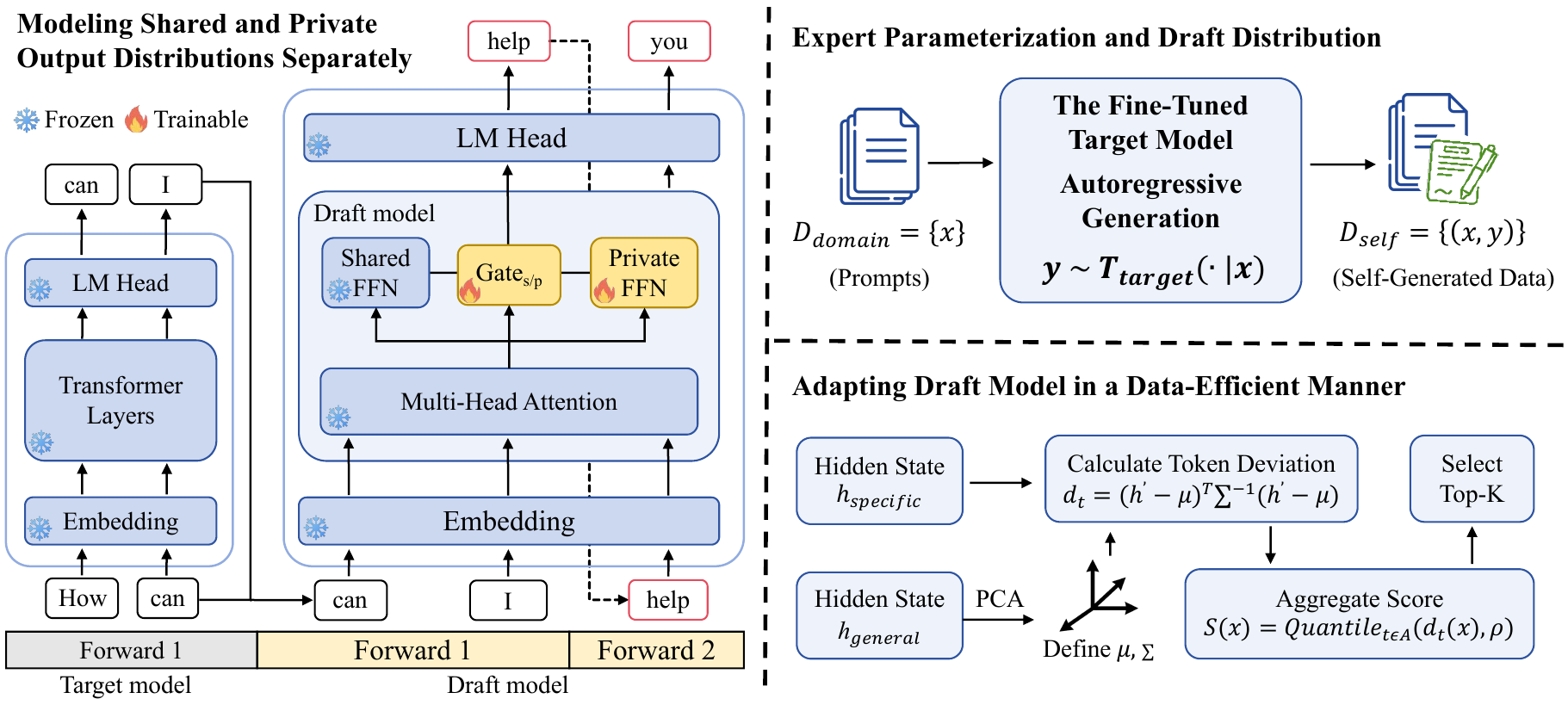}
  \caption{The EDA framework for efficient draft model adaptation, combining shared–private draft decomposition, domain-specific self-generation, and representation-shift–based data selection to restore speculative decoding performance with minimal training cost.}
  \label{fig:framework}
\end{figure*}

This section introduces EDA, a framework designed for efficient adaptation of draft models to align with fine-tuned target models.
EDA integrates a shared--private draft architecture, domain-specific self-generation, and representation-shift-based data selection strategy to efficiently restore speculative decoding performance. Figure~\ref{fig:framework} provides an overview of our EDA framework.

\subsection{Modeling Shared and Private Output Distributions Separately}\label{sec:Separately_Modeling}
As shown in Figure~\ref{fig:transfer_gap}, directly using the draft model $D_{Base}$ trained for the base target model $T_{Base}$ with the fine-tuned target model $T_{Math}$ leads to a significant decrease in average acceptance length.
However, the non-zero acceptance length suggests that their output distributions partially overlap.

We define the overlapping portion of the output distribution shared across different target models as the shared output distribution, while the remaining part is referred to as the target-specific output distribution.
Conventional speculative decoding methods~\cite{li2024eagle,zhou2023distillspec,yan2025dsd} typically train a monolithic draft model to capture the entire target distribution, without differentiating between the shared and target-specific components.
As a result, these methods often require retraining or fine-tuning the entire draft model for each fine-tuned target model, leading to substantial training overhead.
This raises a key question: \textit{Can we efficiently adapt the draft model to align with fine-tuned target models by reusing the shared distribution and focusing only on the target-specific differences?}

To address this question, we decouple the draft model into a shared expert and a private expert, enabling separate modeling of the shared and target-specific output distributions.
When the target model undergoes fine-tuning, the shared expert remains fixed, while only the lightweight private expert is updated. This approach allows the draft model to efficiently adapt to the target-specific output distribution in a parameter-efficient manner.

\paragraph{Shared–Private Gated Draft Architecture.}
Based on the above decomposition, we implement the draft model as a two-expert gated architecture.
Specifically, we replace the feed-forward network (FFN) in the Transformer block with a shared–private gated module, while keeping the attention layers unchanged.
The module comprises a shared expert $E^{\mathrm{s}}$ and a private expert $E^{\mathrm{p}}$, combined via a learnable gating mechanism.
Given the hidden representation $h_t \in \mathbb{R}^{d}$ of the $t$-th token, the outputs of two experts are computed as:
\begin{equation}
\bigl[u_t^{\mathrm{s}},\, u_t^{\mathrm{p}}\bigr]
=
\bigl[E^{\mathrm{s}}(h_t),\, E^{\mathrm{p}}(h_t)\bigr].
\label{eq:spmoe_experts}
\end{equation}
The gate values $g_t^s$ and $g_t^p$ are computed as:
\begin{equation}
\bigl[g_t^{\mathrm{s}},\, g_t^{\mathrm{p}}\bigr]
=
\operatorname{softmax}
\bigl(
[ (w^{\mathrm{s}})^{\top} h_t,\; (w^{\mathrm{p}})^{\top} h_t ]
\bigr),
\label{eq:spmoe_gates}
\end{equation}
where $w^{\mathrm{s}}, w^{\mathrm{p}} \in \mathbb{R}^{d}$ are learnable routing weights.
The softmax function ensures the gate values sum to 1.
The output of the shared–private experts gated module is then computed as:
\begin{equation}
\tilde{h}_t
=
g_t^{\mathrm{s}}\, u_t^{\mathrm{s}}
+
g_t^{\mathrm{p}}\, u_t^{\mathrm{p}}.
\label{eq:spmoe_out}
\end{equation}
This design creates a lightweight gated draft model with two experts, where routing dynamically adjusts the contributions of the shared and private components.

\paragraph{Expert Parameterization and Draft Distribution.}
Having defined the two-expert gated architecture and its routing mechanism, we now specify the parameterization of each expert and describe how the combined hidden representation is mapped to the draft token distribution. Each expert is parameterized as a standard bottleneck MLP:
\begin{equation}
E(h) = U\, \phi(Vh),
\label{eq:mlp}
\end{equation}
where $\phi(\cdot)$ is a nonlinear activation function,
$V \in \mathbb{R}^{m \times d}$ and $U \in \mathbb{R}^{d \times m}$ are the down- and up-projection matrices with $m$ being the intermediate dimension.
The draft model produces token probabilities through:
\begin{equation}
P_{\theta_{d}}(x_t \mid x_{<t}) = \mathrm{softmax}(W_o \tilde{h}_t),
\label{eq:draft_softmax}
\end{equation}
where $W_o \in \mathbb{R}^{|\mathcal{V}|\times d}$ is the output projection head and $|\mathcal{V}|$ denotes the vocabulary size.

\paragraph{Draft Model Initialization and Adaptation.}
The shared expert $E^{\mathrm{s}}$ is initialized using the parameters of a draft model pre-trained on general datasets, such as $D_{Base}$ trained on ShareGPT~\cite{ShareGPT}, to model the shared output distribution.
For adaptation to a fine-tuned target model like $T_{Math}$, the shared expert $E^{\mathrm{s}}$ remains frozen, while only the private expert $E^{\mathrm{p}}$ and the routing parameters $w^{\mathrm{p}}$ and $w^{\mathrm{s}}$ are updated. This allows the private expert to model target-specific output distributions, with the gating mechanism dynamically combining the contributions of both experts.

\subsection{Matching Training and Drafting Objective}\label{self-generate}
We further improve the average acceptance length of our EDA by addressing the mismatch between the draft model's training objectives and speculative decoding objectives.
Specifically, consider two token sequences: $x = \{x_{<t}, \cdots, x_n\}$ sampled from the data distribution $\mathcal{D}$, and $x' = \{x_{<t}, \cdots, x'_n\}$ generated by the target model via autoregressive sampling conditioned on the same prefix $x_{<t}$. Since the target model's prediction distribution may not perfectly align with $\mathcal{D}$, $x_n$ and $x'_n$ can differ.

Traditionally, the draft model is trained on datasets like the first token sequence $x$ using Eq.~(\ref{eq:draft_training_loss})~\cite{li2024eagle}. However, during speculative decoding, it predicts the next token in the second token sequence $x'$ generated by the target model. This mismatch between the training and drafting objectives leads to suboptimal average acceptance length.

In domain-specific fine-tuning scenarios, such as medicine, law, or code, proprietary datasets used for fine-tuning the target model are often inaccessible due to privacy or ownership restrictions. As a result, the draft model must be trained on publicly available datasets, which may significantly differ from the proprietary data distribution, further widening the gap between the draft model's training and speculative decoding objectives, ultimately reducing the average acceptance length.

To address this mismatch, we use a domain-specific self-generation strategy, in which the fine-tuned target model provides supervision during the draft model training stage. Given a domain-specific dataset $\mathcal{D}_{\mathrm{domain}}=\{x^{i}\}$, for each $x^{i}$ we sample the target model's completions via conditional autoregressive generation:
\begin{equation}
y^{i} \sim P_{\theta_t}(\,\cdot \mid x^{i}\,).
\label{eq:selfgen_target}
\end{equation}
This procedure produces the training set generated by the target model itself:
\begin{equation}
\mathcal{D}_{\mathrm{self}} = \{(x^{i},\, y^{i})\}.
\label{eq:selfgen_dataset}
\end{equation}
We then optimize the draft model on $\mathcal{D}_{\mathrm{self}}$ via:
\begin{equation}
\begin{split}
  \theta_{d} =  \arg\min_{\theta_d} \, & \mathbb{E}_{x \sim \mathcal{D}_{\mathrm{self}}} \sum_{t} \Bigl[
    L_{\mathrm{reg}} + \mathrm{CE}\bigl(x_t, P_{\theta_{d}}(\cdot \mid x_{<t})\bigr)
  \Bigr].
\end{split}
\end{equation}
This objective performs trajectory-level alignment on target-generated tokens rather than full distribution distillation.
By matching training and drafting objectives, our EDA achieves a higher average acceptance length during speculative decoding, as shown in our experiments.

\subsection{Adapting Draft Model in a Data-Efficient Manner}
Training draft models in the entire domain-specific datasets can still incur significant costs, even when optimization is restricted to the private expert.
Thus, we further improve draft model adaptation efficiency from the data perspective.

As discussed in Sec.~\ref{sec:Separately_Modeling}, the shared expert captures the shared output distribution, while the private expert focuses on modeling the target-specific output distribution.
To achieve efficient draft model adaptation, it is crucial to prioritize samples that exhibit significant deviations from the shared output distribution, as these provide the most valuable information for adapting the private expert, while filtering out samples that are already well-represented.
Building on the observation that specialization often manifests as structured shifts in the representation space~\cite{lee2018samples}, we introduce a training-free sample value metric based solely on the hidden states generated by the target model during self-generation.
This metric is then used to identify and select a subset of the training data that is most beneficial for refining the private expert.

Given a self-generated sample $x \sim \mathcal{D}_{\text{self}}$ of length $L$, let $\mathbf{H} = \{h_t\}_{t=1}^{L}$ denote the sequence of hidden representations produced by the target model. We define $\mathcal{A} \subset \{1, \dots, L\}$ as the set of indices corresponding to the answer tokens.

We first apply Principal Component Analysis (PCA) to reduce the dimensionality of the answer token's hidden representations as:
\begin{equation}
h'_t \leftarrow \mathrm{PCA}(h_t), \quad \forall t
\in \mathcal{A},
\label{eq:pca}
\end{equation}
This reduction ensures that only the most relevant features of the data are retained, improving both computational efficiency and stability for subsequent statistical calculations.

After reducing dimensionality, we calculate the mean $\mu$ and covariance matrix $\Sigma$ of the reduced representations $h'_t$ using a general-generation dataset $\mathcal{D}_{\mathrm{general}}$:
\begin{equation}
\mu = \mathbb{E}_{h'\sim \mathcal{D}_{\mathrm{general}}}[h'],
\quad
\Sigma = \mathrm{Cov}_{h'\sim \mathcal{D}_{\mathrm{general}}}(h').
\label{eq:ref_stats}
\end{equation}
Here, $\mathcal{D}_{\mathrm{general}}$ is the 68k ShareGPT samples used for base draft training; PCA, $\mu$, and $\Sigma$ are fitted once on the final-layer answer-token hidden states of $T_{Base}$, and each fine-tuned target's representations are projected into this fixed space.
We then compute the Mahalanobis score~\cite{Mahalanobis1936OnTG} $d_t(x)$ to quantify each answer-token representation's deviation from the reference distribution $\mathcal{D}_{\mathrm{general}}$.
The score is computed on the PCA-reduced representations $h'^{(x)}_{t}$. Specifically, for each \( t \in \mathcal{A} \), we define
\begin{equation}
d_t(x) = \big(h'^{(x)}_{t} - \mu\big)^\top \Sigma^{-1} \big(h'^{(x)}_{t} - \mu\big).
\label{eq:mahalanobis}
\end{equation}
Since domain- and task-specific behaviors often concentrate in a small number of critical steps or local segments, we aggregate token-level deviations with a tail-sensitive statistic to obtain a sample-level value score $s(x)$. In particular, we use the high-quantile aggregation
\begin{equation}
s(x) = \mathrm{Quantile}_{t\in \mathcal{A}}\!\big(d_t(x), \rho\big),
\label{eq:quantile_score}
\end{equation}
where $\rho$ controls the quantile level.

Given the whole dataset $\mathcal{D}_{\mathrm{self}}$, we rank samples by $s(x)$ and select the subset as:
\begin{equation}
\mathcal{D}_{\mathrm{compact}} =
\bigl\{x_i \in \mathcal{D}_{\mathrm{self}} \mid s(x_i)\ \text{ranks among the top}\ \lfloor \alpha |\mathcal{D}_{\mathrm{self}}| \rfloor\ \text{scores}\bigr\},
\label{eq:topk}
\end{equation}
where $\alpha \in (0,1]$ denotes the selection ratio. The selected subset is used to update the private expert while keeping the shared expert frozen. This strategy shifts training effort from representation regions already covered by the shared expert to answer segments that exhibit larger deviations from the general reference, improving adaptation efficiency under a limited sample budget. Importantly, the selection process incurs no extra forward passes or auxiliary components, as the hidden representations $h_t$ are directly reused from the self-generation stage.

\begin{table}[!t]
  \centering
  \caption{Main results on five math tasks (GSM8K, AIME\_2024, SVAMP, H-MATH, MathQA) on Qwen2.5-Math-7B target model~\cite{yang2024qwen25math}.
  We report average acceptance length $\tau$ and decoding speedup, where Speed. denotes decoding speedup.}
  \label{tab:main_math_results}

  \footnotesize
  \setlength{\tabcolsep}{5.2pt}

  \begin{tabular}{lcccccccccccc}
    \toprule
    & \multicolumn{2}{c}{\textbf{GSM8K}} & \multicolumn{2}{c}{\textbf{AIME\_2024}} & \multicolumn{2}{c}{\textbf{SVAMP}} &
      \multicolumn{2}{c}{\textbf{H-MATH}} & \multicolumn{2}{c}{\textbf{MathQA}} & \multicolumn{2}{c}{\textbf{Avg}} \\
    \midrule
    Method
    & $\tau$ & Speed.
    & $\tau$ & Speed.
    & $\tau$ & Speed.
    & $\tau$ & Speed.
    & $\tau$ & Speed.
    & $\tau$ & Speed. \\
    \midrule
    \multicolumn{13}{c}{Temperature $T=0$} \\
    \midrule
    Training-Free & 1.17 & 0.85$\times$ & 1.18 & 0.76$\times$ & 1.19 & 0.85$\times$ & 1.20 & 0.82$\times$ & 1.23 & 0.91$\times$ & 1.19 & 0.84$\times$ \\
    Full-FT       & 4.37 & 2.88$\times$ & 5.05 & 3.14$\times$ & 4.40 & 2.91$\times$ & 4.98 & 3.23$\times$ & 4.66 & 3.19$\times$ & 4.69 & 3.07$\times$ \\
    LoRA          & 4.32 & 2.84$\times$ & 4.90 & 2.93$\times$ & 4.36 & 2.75$\times$ & 4.92 & 3.18$\times$ & 4.55 & 3.00$\times$ & 4.61 & 2.94$\times$ \\
    \textbf{EDA} & \textbf{4.79} & \textbf{3.06$\times$} & \textbf{5.41} & \textbf{3.20$\times$} & \textbf{4.96} & \textbf{3.09$\times$}
                        & \textbf{5.60} & \textbf{3.59$\times$} & \textbf{5.16} & \textbf{3.43$\times$} & \textbf{5.18} & \textbf{3.27$\times$} \\
    \midrule
    \multicolumn{13}{c}{Temperature $T=1$} \\
    \midrule
    Training-Free & 1.02 & 0.65$\times$ & 1.15 & 0.72$\times$ & 1.08 & 0.68$\times$ & 1.21 & 0.75$\times$ & 1.04 & 0.66$\times$ & 1.10 & 0.69$\times$ \\
    Full-FT       & 2.65 & 1.72$\times$ & 2.89 & 1.84$\times$ & 2.74 & 1.76$\times$ & 3.12 & 1.95$\times$ & 3.01 & 1.90$\times$ & 2.88 & 1.83$\times$ \\
    LoRA          & 2.60 & 1.70$\times$ & 2.80 & 1.78$\times$ & 2.70 & 1.74$\times$ & 3.05 & 1.92$\times$ & 2.93 & 1.86$\times$ & 2.82 & 1.80$\times$ \\
    \textbf{EDA} & \textbf{3.15} & \textbf{1.92$\times$} & \textbf{3.38} & \textbf{2.05$\times$} & \textbf{3.26} & \textbf{1.98$\times$}
                        & \textbf{3.75} & \textbf{2.24$\times$} & \textbf{3.52} & \textbf{2.12$\times$} & \textbf{3.41} & \textbf{2.06$\times$} \\
    \bottomrule
  \end{tabular}
\end{table}

\begin{table}[!t]
  \centering
  \caption{Main results on five code tasks (HumanEval, APPS, BigCodeBench, HumanEval+, MBPP) on Qwen2.5-Coder-7B target model~\cite{hui2024qwen2.5coder}.
  }
  \label{tab:main_code_results}

  \footnotesize
  \setlength{\tabcolsep}{5.2pt}

  \begin{tabular}{lcccccccccccc}
    \toprule
    & \multicolumn{2}{c}{\textbf{HumanEval}} & \multicolumn{2}{c}{\textbf{APPS}} & \multicolumn{2}{c}{\textbf{BigCodeBench}} &
      \multicolumn{2}{c}{\textbf{HumanEval+}} & \multicolumn{2}{c}{\textbf{MBPP}} & \multicolumn{2}{c}{\textbf{Avg}} \\
    \midrule
    Method
    & $\tau$ & Speed.
    & $\tau$ & Speed.
    & $\tau$ & Speed.
    & $\tau$ & Speed.
    & $\tau$ & Speed.
    & $\tau$ & Speed. \\
    \midrule
    \multicolumn{13}{c}{Temperature $T=0$} \\
    \midrule
    Training-Free & 1.75 & 1.21$\times$ & 1.69 & 1.14$\times$ & 1.74 & 1.24$\times$ & 1.74 & 1.08$\times$ & 1.85 & 1.21$\times$ & 1.75 & 1.18$\times$ \\
    Full-FT       & 4.79 & 3.10$\times$ & 4.87 & 2.91$\times$ & 3.66 & 2.38$\times$ & 4.70 & 2.62$\times$ & 4.93 & 2.82$\times$ & 4.59 & 2.77$\times$ \\
    LoRA          & 4.70 & 3.05$\times$ & 4.80 & 2.88$\times$ & 3.60 & 2.35$\times$ & 4.62 & 2.58$\times$ & 4.78 & 2.76$\times$ & 4.50 & 2.72$\times$ \\
    \textbf{EDA} & \textbf{5.35} & \textbf{3.36$\times$} & \textbf{5.65} & \textbf{3.34$\times$} & \textbf{4.18} & \textbf{2.67$\times$}
                        & \textbf{5.31} & \textbf{2.98$\times$} & \textbf{5.43} & \textbf{3.18$\times$} & \textbf{5.18} & \textbf{3.11$\times$} \\
    \midrule
    \multicolumn{13}{c}{Temperature $T=1$} \\
    \midrule
    Training-Free & 1.28 & 0.78$\times$ & 1.35 & 0.81$\times$ & 1.22 & 0.75$\times$ & 1.41 & 0.85$\times$ & 1.30 & 0.79$\times$ & 1.31 & 0.80$\times$ \\
    Full-FT       & 3.72 & 2.00$\times$ & 4.05 & 2.13$\times$ & 3.35 & 1.80$\times$ & 3.88 & 2.05$\times$ & 4.32 & 2.23$\times$ & 3.86 & 2.04$\times$ \\
    LoRA          & 3.60 & 1.95$\times$ & 3.95 & 2.08$\times$ & 3.30 & 1.78$\times$ & 3.80 & 2.00$\times$ & 4.25 & 2.18$\times$ & 3.78 & 2.00$\times$ \\
    \textbf{EDA} & \textbf{4.38} & \textbf{2.28$\times$} & \textbf{4.62} & \textbf{2.38$\times$} & \textbf{3.95} & \textbf{2.06$\times$}
                        & \textbf{4.55} & \textbf{2.32$\times$} & \textbf{4.92} & \textbf{2.48$\times$} & \textbf{4.48} & \textbf{2.30$\times$} \\
    \bottomrule
  \end{tabular}
\end{table}

\begin{table}[!t]
  \centering
  \caption{Main results on five medical tasks (MedMCQA, MedQA-USMLE, PubMedQA, MedQA, MMLU) on Meditron3-Qwen2.5-7B target model~\cite{meditron3_qwen25}.
  }
  \label{tab:main_medical_results}

  \footnotesize
  \setlength{\tabcolsep}{4.8pt}

  \begin{tabular}{lcccccccccccc}
    \toprule
    & \multicolumn{2}{c}{\textbf{MedMCQA}} & \multicolumn{2}{c}{\textbf{MedQA-USMLE}} & \multicolumn{2}{c}{\textbf{PubMedQA}} &
      \multicolumn{2}{c}{\textbf{MedQA}} & \multicolumn{2}{c}{\textbf{MMLU}} & \multicolumn{2}{c}{\textbf{Avg}} \\
    \midrule
    Method
    & $\tau$ & Speed.
    & $\tau$ & Speed.
    & $\tau$ & Speed.
    & $\tau$ & Speed.
    & $\tau$ & Speed.
    & $\tau$ & Speed. \\
    \midrule
    \multicolumn{13}{c}{Temperature $T=0$} \\
    \midrule
    Training-Free & 2.20 & 1.47$\times$ & 2.17 & 1.40$\times$ & 1.92 & 1.29$\times$ & 2.14 & 1.29$\times$ & 2.33 & 1.49$\times$ & 2.15 & 1.39$\times$ \\
    Full-FT       & 3.97 & 2.35$\times$ & 3.97 & 2.28$\times$ & 3.86 & 2.34$\times$ & 3.97 & 2.35$\times$ & 3.67 & 3.03$\times$ & 3.89 & 2.47$\times$ \\
    LoRA          & 3.85 & 2.30$\times$ & 3.90 & 2.25$\times$ & 3.78 & 2.18$\times$ & 3.92 & 2.28$\times$ & 3.50 & 2.80$\times$ & 3.79 & 2.36$\times$ \\
    \textbf{EDA} & \textbf{4.30} & \textbf{2.55$\times$} & \textbf{4.34} & \textbf{2.55$\times$} & \textbf{4.03} & \textbf{2.40$\times$}
                        & \textbf{4.32} & \textbf{2.51$\times$} & \textbf{4.06} & \textbf{3.29$\times$} & \textbf{4.21} & \textbf{2.66$\times$} \\
    \midrule
    \multicolumn{13}{c}{Temperature $T=1$} \\
    \midrule
    Training-Free & 1.42 & 0.88$\times$ & 1.55 & 0.95$\times$ & 1.38 & 0.82$\times$ & 1.62 & 1.02$\times$ & 1.48 & 0.92$\times$ & 1.49 & 0.92$\times$ \\
    Full-FT       & 2.42 & 1.42$\times$ & 2.65 & 1.52$\times$ & 2.28 & 1.35$\times$ & 2.72 & 1.56$\times$ & 2.55 & 1.48$\times$ & 2.52 & 1.47$\times$ \\
    LoRA          & 2.35 & 1.38$\times$ & 2.60 & 1.50$\times$ & 2.25 & 1.33$\times$ & 2.68 & 1.52$\times$ & 2.50 & 1.46$\times$ & 2.48 & 1.44$\times$ \\
    \textbf{EDA} & \textbf{2.85} & \textbf{1.62$\times$} & \textbf{3.12} & \textbf{1.75$\times$} & \textbf{2.72} & \textbf{1.56$\times$}
                        & \textbf{3.25} & \textbf{1.82$\times$} & \textbf{3.08} & \textbf{1.72$\times$} & \textbf{3.00} & \textbf{1.69$\times$} \\
    \bottomrule
  \end{tabular}
\end{table}

\section{Experiments}
\subsection{Experimental Configurations}
\paragraph{Dataset.}
We evaluate EDA across three downstream domains, including mathematical reasoning, code generation, and medical question answering.
For each domain, we use five representative benchmarks and follow the standard evaluation protocols of each benchmark to construct prompts.
Detailed descriptions of the evaluation benchmarks and training datasets are provided in Appendix~\ref{app:data}.

\paragraph{Model Setup.}
We evaluate EDA in a practical setting where a general-purpose base model is fine-tuned into multiple domain-specific target models.
To avoid the prohibitive cost of retraining target models ourselves, we use off-the-shelf models from the Qwen2.5 family~\cite{qwen2025qwen2.5}: Qwen2.5-7B as the base target, and Qwen2.5-Math-7B~\cite{yang2024qwen25math}, Qwen2.5-Coder-7B~\cite{hui2024qwen2.5coder}, and Meditron3-Qwen2.5-7B~\cite{meditron3_qwen25} as domain-specific fine-tuned targets.
This setup preserves architectural consistency while simulating real-world target-model specialization.
Additional experiments on another model architecture are provided in Appendix~\ref{app:llama}. The draft model architecture and implementation details are described in Appendix~\ref{app:impl}.

\paragraph{Baseline.} We compare EDA with three representative baselines that share the same pre-trained backbone model but differ in their adaptation strategies, training cost, and parameter efficiency.
(1) \textbf{Training-Free} directly uses a pre-trained model as the draft model without any additional training or adaptation to the fine-tuned target model.
(2) \textbf{Full Fine-Tuning (Full-FT)} adapts the same backbone model by updating all parameters on the available adaptation set, representing the strongest but most expensive transfer baseline.
(3) \textbf{Low-Rank Adaptation (LoRA)}~\cite{hu2022lora} performs parameter-efficient transfer by training only low-rank update matrices while freezing the backbone. We match the number of trainable parameters to EDA to ensure a fair comparison in terms of parameter efficiency.
For Full-FT and LoRA, we use the same publicly available domain adaptation datasets under an identical data budget. For EDA, we use the same domain data as prompts to construct the target-model self-generated adaptation set described in Section~\ref{self-generate}, and apply the proposed representation-shift-based data selection strategy to retain 50\% of the adaptation samples.

\paragraph{Evaluation Metrics.}
We evaluate speculative decoding using average acceptance length $\tau$ and decoding speedup. Average acceptance length, formally defined in Appendix~\ref{sec:avg_acceptance_length}, measures the expected number of accepted draft tokens before rejection, and speedup is the latency ratio between standard autoregressive decoding and speculative decoding. Since speculative decoding is lossless, we focus on efficiency metrics and do not separately report task accuracy.

\subsection{Main Results}

Tables~\ref{tab:main_math_results}--\ref{tab:main_medical_results} summarize the main results across math, code, and medical benchmarks under both greedy ($T=0$) and stochastic ($T=1$) decoding. Across all three domains, EDA consistently improves both average acceptance length $\tau$ and decoding speedup over Training-Free, Full-FT, and LoRA baselines.

The Training-Free baseline performs poorly after target-model fine-tuning, confirming that directly reusing the original draft model cannot handle the distribution shifts introduced by domain specialization. Compared with Full-FT and parameter-matched LoRA, EDA achieves stronger speculative decoding performance in most settings while updating only a lightweight target-specific component and using a reduced adaptation set. These results show that combining shared--private draft adaptation, target-model self-generation, and representation-shift-based data selection effectively restores draft--target alignment under limited adaptation cost. Unless otherwise specified, subsequent evaluations follow the math transfer setting with $T=0$ on a single H200 GPU.

\subsection{Component-wise Ablation Study}

\begin{table}[!t]
  \centering
  \caption{Component-wise ablation on Qwen2.5-Math-7B. We report average acceptance length $\tau$ and decoding speedup on representative math benchmarks.}
  \label{tab:component_ablation}
  \footnotesize
  \setlength{\tabcolsep}{4pt}
  \begin{tabular}{lccccc}
    \toprule
    Method & Trainable Params & Samples & GSM8K & AIME & MathQA \\
    \midrule
    Full-FT & 462MB & 68k & 4.37 / 2.88$\times$ & 5.05 / 3.14$\times$ & 4.66 / 3.19$\times$ \\
    + Shared/Private FFN & 127MB & 68k & 4.40 / 2.89$\times$ & 5.02 / 3.12$\times$ & 4.70 / 3.20$\times$ \\
    + Self-generated Data & 127MB & 68k & 4.83 / 3.08$\times$ & 5.46 / 3.23$\times$ & 5.20 / 3.46$\times$ \\
    + Data Selection & 127MB & 34k & 4.79 / 3.06$\times$ & 5.41 / 3.20$\times$ & 5.16 / 3.43$\times$ \\
    \bottomrule
  \end{tabular}
  \vspace{-4pt}
\end{table}

To understand how each component contributes to EDA, we conduct a cumulative ablation on the Qwen2.5-Math-7B target model. As shown in Table~\ref{tab:component_ablation}, the shared/private FFN architecture matches the performance of Full-FT while reducing the trainable parameters from 462MB to 127MB. Replacing the original adaptation data with target-model self-generated data provides the largest improvement, confirming that aligning the training objective with the speculative decoding trajectory is crucial for draft adaptation. Finally, the representation-shift-based data selection reduces the adaptation samples by 50\% with only negligible performance loss. These results show that the three components contribute in complementary ways: the architecture improves parameter efficiency, self-generation improves target alignment, and data selection improves data efficiency.

\subsection{Effectiveness of Separately Modeling Shared and Private Distributions}

\begin{table*}[!t]
    \caption{
    Comparison of adaptation structures and training overhead.
    The left part reports performance on math and general benchmarks, where each cell denotes $\tau$ / speedup.
    The right part reports the adaptation overhead on math transfer, following the same experimental setup as in Sec.~\ref{data_selection}.
    }
  \label{tab:arch_overhead}
  \vspace{4pt}
  \footnotesize

  \noindent
  \begin{minipage}[t]{0.595\textwidth}
    \centering

    \resizebox{\linewidth}{!}{
    \begin{tabular}{cccccc}
      \toprule
      & & \multicolumn{2}{c}{\textbf{Math Benchmarks}}
      & \multicolumn{2}{c}{\textbf{General Benchmarks}} \\
      \cmidrule(lr){3-4} \cmidrule(lr){5-6}
      Method & Params
      & GSM8K & MathQA
      & Alpaca & MT-Bench \\
      \midrule
      Full-FT
      & 462MB
      & \textbf{4.82 / 3.08$\times$}
      & \textbf{5.20 / 3.46$\times$}
      & 5.17 / 3.02$\times$
      & 4.19 / 2.67$\times$ \\

      LoRA
      & \textbf{127MB}
      & 4.66 / 2.99$\times$
      & 5.07 / 3.38$\times$
      & 4.89 / 2.76$\times$
      & 4.04 / 2.40$\times$ \\

      EDA
      & \textbf{127MB}
      & 4.79 / 3.06$\times$
      & 5.16 / 3.43$\times$
      & \textbf{5.21 / 3.04$\times$}
      & \textbf{4.22 / 2.68$\times$} \\
      \bottomrule
    \end{tabular}
    }
  \end{minipage}
  \hspace{0.01\textwidth}
  \begin{minipage}[t]{0.38\textwidth}
    \centering

    \footnotesize
    \setlength{\tabcolsep}{3pt}
    \renewcommand{\arraystretch}{1.08}
    \begin{tabular}{ccc}
      \toprule
      Metric & Re-Train & EDA \\
      \midrule
      Trainable Size & 462 MB & \textbf{127 MB} (27.5\%) \\
      Training Time & 5.1 h & \textbf{2.0 h} (39.2\%) \\
      $\tau$ & 4.22 & \textbf{4.79} (+13.5\%) \\
      \bottomrule
    \end{tabular}
  \end{minipage}
\end{table*}

We compare EDA with Full-FT and parameter-matched LoRA, all trained on the same self-generated and selected data. As shown in the left of Table~\ref{tab:arch_overhead}, EDA matches Full-FT with only 27.5\% of its trainable parameters and outperforms LoRA, with a larger margin on the general-domain benchmarks (Alpaca and MT-Bench), suggesting that LoRA interferes with reusable drafting capability while the frozen shared expert preserves it.

\begin{table}[!t]
  \centering
  \caption{Routing analysis and branch ablation of the Math-adapted EDA draft model on answer tokens. Entropy is the base-2 routing entropy (1 for uniform routing). Shared-/Private-only zero out the other expert; entries report CE to the target distribution ($\downarrow$) / top-1 agreement with the target ($\uparrow$).}
  \label{tab:routing_analysis}
  \footnotesize
  \setlength{\tabcolsep}{4pt}
  \begin{tabular}{lccccc}
    \toprule
    & \multicolumn{3}{c}{\textbf{Routing Analysis}} & \multicolumn{2}{c}{\textbf{Branch Ablation (CE / Top-1)}} \\
    \cmidrule(lr){2-4} \cmidrule(lr){5-6}
    Domain & Shared Gate & Private Gate & Entropy & Shared-only & Private-only \\
    \midrule
    General / Alpaca & \textbf{58.64\%} & 41.36\% & 0.98 & \textbf{5.63 / 25.54\%} & 8.71 / 7.11\% \\
    Math / GSM8K & 29.57\% & \textbf{70.43\%} & 0.88 & 10.18 / 11.76\% & \textbf{4.60 / 39.96\%} \\
    \bottomrule
  \end{tabular}
\end{table}

To verify that the experts learn their intended roles, we analyze the Math-adapted EDA checkpoint on Alpaca and GSM8K samples (Table~\ref{tab:routing_analysis}). The router favors the shared expert on Alpaca and the private expert on GSM8K, and branch ablation shows the same reversal: shared-only aligns better with the target model on Alpaca, whereas private-only aligns better on GSM8K. This confirms that the shared expert retains general drafting behavior, while the private expert captures the domain fine-tuning shift.

\subsection{Effectiveness of Adapting Draft Model in a Data-Efficient Manner}
\label{data_selection}

We evaluate the effect of our representation-shift-based data selection under different data budgets. As shown in Figure~\ref{fig:selection_metrics}, the proposed selection strategy achieves near full-data performance with only 50\% of the adaptation data. This indicates that representation shift is an effective proxy for identifying samples that are more useful for adapting the private expert. Detailed comparisons with alternative selection metrics and additional analysis are provided in Appendix~\ref{app:data_selection}.

\subsection{Adaptation Overhead Analysis}

As shown in the right part of Table~\ref{tab:arch_overhead}, on the math transfer task from Qwen2.5-7B to Qwen2.5-Math-7B, EDA substantially outperforms training a draft model from scratch in both efficiency and performance. Re-Train updates 462 MB of parameters and takes 5.1 hours, whereas EDA adapts only 127 MB of parameters (27.5\%) and finishes in 2.0 hours (39.2\%). Despite the reduced training cost, EDA achieves a higher average acceptance length, improving $\tau$ from 4.22 to 4.79. These results demonstrate that EDA delivers better alignment while markedly reducing adaptation time.

\subsection{Scaling to Larger Target Models}

\begin{table}[!t]
  \centering
  \caption{Results on five math tasks with a larger target model, OpenMath-Nemotron-14B, which is fine-tuned from Qwen2.5-14B. We report average acceptance length $\tau$ and decoding speedup.}
  \label{tab:main_math_14b_results}

  \footnotesize
  \setlength{\tabcolsep}{5.2pt}

  \begin{tabular}{lcccccccccccc}
    \toprule
    & \multicolumn{2}{c}{\textbf{GSM8K}} & \multicolumn{2}{c}{\textbf{AIME\_2024}} & \multicolumn{2}{c}{\textbf{SVAMP}} &
      \multicolumn{2}{c}{\textbf{H-MATH}} & \multicolumn{2}{c}{\textbf{MathQA}} & \multicolumn{2}{c}{\textbf{Avg}} \\
    \midrule
    Method
    & $\tau$ & Speed.
    & $\tau$ & Speed.
    & $\tau$ & Speed.
    & $\tau$ & Speed.
    & $\tau$ & Speed.
    & $\tau$ & Speed. \\
    \midrule
    \multicolumn{13}{c}{Temperature $T=0$} \\
    \midrule
    Training-Free & 1.72 & 1.25$\times$ & 1.80 & 1.31$\times$ & 1.69 & 1.22$\times$ & 1.91 & 1.34$\times$ & 1.84 & 1.28$\times$ & 1.79 & 1.28$\times$ \\
    Full-FT       & 4.31 & 2.95$\times$ & 4.52 & 3.08$\times$ & 4.35 & 3.04$\times$ & 4.72 & 3.13$\times$ & 4.73 & 3.20$\times$ & 4.53 & 3.08$\times$ \\
    LoRA          & 4.24 & 2.88$\times$ & 4.43 & 3.01$\times$ & 4.26 & 2.96$\times$ & 4.66 & 3.07$\times$ & 4.61 & 3.14$\times$ & 4.44 & 3.01$\times$ \\
    \textbf{EDA} & \textbf{4.69} & \textbf{3.14$\times$} & \textbf{5.02} & \textbf{3.39$\times$} & \textbf{4.68} & \textbf{3.17$\times$}
                        & \textbf{5.30} & \textbf{3.48$\times$} & \textbf{5.03} & \textbf{3.44$\times$} & \textbf{4.94} & \textbf{3.32$\times$} \\
    \midrule
    \multicolumn{13}{c}{Temperature $T=1$} \\
    \midrule
    Training-Free & 1.58 & 1.14$\times$ & 1.66 & 1.20$\times$ & 1.61 & 1.16$\times$ & 1.74 & 1.25$\times$ & 1.70 & 1.19$\times$ & 1.66 & 1.19$\times$ \\
    Full-FT       & 4.10 & 2.74$\times$ & 4.37 & 2.80$\times$ & 4.01 & 2.65$\times$ & 4.59 & 2.96$\times$ & 4.44 & 2.89$\times$ & 4.30 & 2.81$\times$ \\
    LoRA          & 4.02 & 2.68$\times$ & 4.28 & 2.73$\times$ & 3.95 & 2.58$\times$ & 4.52 & 2.89$\times$ & 4.35 & 2.82$\times$ & 4.22 & 2.74$\times$ \\
    \textbf{EDA} & \textbf{4.51} & \textbf{2.95$\times$} & \textbf{4.85} & \textbf{3.15$\times$} & \textbf{4.56} & \textbf{2.88$\times$}
                        & \textbf{5.16} & \textbf{3.36$\times$} & \textbf{4.83} & \textbf{3.18$\times$} & \textbf{4.78} & \textbf{3.10$\times$} \\
    \bottomrule
  \end{tabular}
\end{table}

To examine whether EDA scales to larger target models, we adapt the draft model of Qwen2.5-14B to OpenMath-Nemotron-14B, a math-specialized model fine-tuned from Qwen2.5-14B. As shown in Table~\ref{tab:main_math_14b_results}, EDA achieves the highest $\tau$ and speedup on all five benchmarks under both temperatures, improving the average $\tau$ over Full-FT from 4.53 to 4.94 at $T=0$ and from 4.30 to 4.78 at $T=1$.

\vspace{-4pt}

\begin{figure}[!t]
  \centering
  \includegraphics[width=0.750\linewidth]{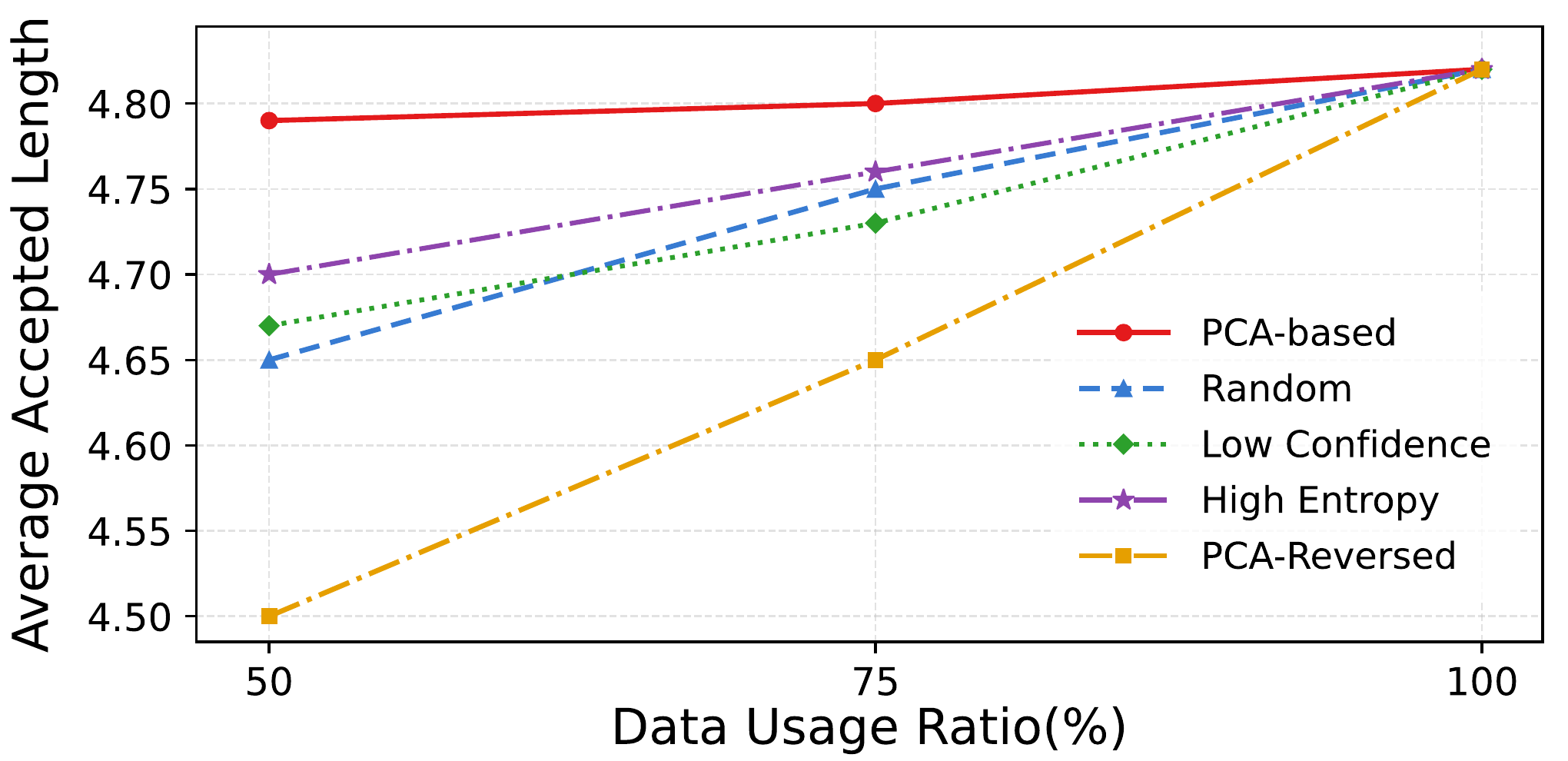}
\caption{
Comparison of data selection strategies under varying data budgets.
Draft adaptation is evaluated from Qwen2.5-7B to Qwen2.5-Math-7B on GSM8K using average acceptance length $\tau$.
}
\label{fig:selection_metrics}
\vspace{-4pt}
\end{figure}

\section{Conclusion}
In this paper, we propose EDA, a parameter- and data-efficient framework for adapting draft models to fine-tuned target models in speculative decoding.
By decoupling the draft model into shared and private components, EDA reuses shared knowledge while updating only lightweight target-specific parameters.
Combined with domain-specific self-generation and representation-shift–based data selection, EDA effectively aligns draft models with evolving target distributions under limited training budgets.
Extensive experiments across math, code, and medical domains show that EDA restores average acceptance length and decoding speedup while reducing adaptation cost compared to full retraining.
Our results highlight the practicality of efficient draft adaptation and suggest a promising direction for scalable speculative decoding in continually evolving LLM systems.

\newpage
\begin{ack}

This work was supported by the New Generation Artificial Intelligence-National Science and Technology Major Project (No. 2025ZD0122703), the National Science Fund for Distinguished Young Scholars (No. 62525605), the National Natural Science Foundation of China (No. U25B2066, No. U22B2051, No. 62506313), and the Xiamen Municipal Science and Technology Bureau, China (3502ZC-QXT2024009).

\end{ack}

\bibliographystyle{plainnat}   %
\bibliography{references}            %

\appendix

\section{Implementation Details}
\label{app:impl}

\paragraph{Hyperparameters.}
All methods are trained with identical optimization settings for fair comparison. We use a batch size of 16, a learning rate of $4\times10^{-5}$, and train for 20 epochs. The maximum training sequence length is 2048 tokens, and the loss is computed only on assistant responses. For LoRA, we apply low-rank updates to all layers involved in draft adaptation and match its trainable parameter size with EDA. In EDA, the shared expert is initialized from the first-stage general draft model and frozen during adaptation, while the private expert and routers are randomly initialized and updated. For representation-shift-based selection, we reduce hidden representations to a 256-dimensional PCA space, add diagonal regularization to the covariance matrix, and select the top $\alpha$ fraction of samples; we set $\alpha=0.5$ in the main experiments. During speculative decoding, we follow the default EAGLE-2 inference configuration with a draft-tree budget of 60 tokens, maximum tree depth of 5, and 10 retained candidates per step. We use batch size 1 and a maximum generation length of 512 tokens. We evaluate decoding under $T=0$ and $T=1$. All training experiments are conducted on 4 NVIDIA H200 GPUs, while inference latency and decoding speedup are measured on a single NVIDIA H200 GPU.

\paragraph{Draft Model Architecture.}
Our implementation starts from an EAGLE2-style draft model architecture and replaces the standard FFN module in each decoder layer with the proposed shared-private MoE module. The self-attention layers, layer normalization, embedding layer, and output prediction head follow the original draft model design. In each shared-private MoE block, a shared-private softmax router computes the gating weights for the shared and private experts jointly. During adaptation, the shared experts are initialized from the base draft model and kept frozen, while the private experts and routing parameters are updated for the fine-tuned target model. This implementation follows the design principle in Section~\ref{sec:Separately_Modeling}: preserving reusable drafting behavior in the shared component while assigning lightweight trainable capacity to target-specific shifts.

\paragraph{Average Acceptance Length.}
\label{sec:avg_acceptance_length}
Average acceptance length $\tau$ measures the expected number of consecutive draft tokens accepted by the target model before a rejection occurs:
\begin{equation}
\tau
=\mathbb{E}_{x_{<t} \sim \mathcal{D}_{\text{test}}}\!\left[\sum_{i=1}^{K}\prod_{j=0}^{i-1} a_{t+j}\right],
\end{equation}
where $K$ is the maximum number of candidate tokens proposed, $a_{t+j} \in \{0,1\}$ indicates whether the draft token $\hat{x}_{t+j}$ is accepted ($a_{t+j} = 1$), and $\mathcal{D}_{\text{test}}$ denotes the test data distribution. A higher $\tau$ indicates better alignment between the draft and target distributions, enabling more tokens to be generated per target-model forward pass and thereby enhancing inference efficiency.

\section{Datasets}
\label{app:data}

\subsection{Evaluation Benchmarks}

We evaluate EDA on three downstream domains: mathematical reasoning, code generation, and medical question answering.

For mathematical reasoning, we use GSM8K~\cite{cobbe2021gsm8k}, AIME2024~\cite{huggingfaceh4_aime_2024}, SVAMP~\cite{patel2021SVAMP}, H-MATH~\cite{hendrycks2021Hendrycks}, and MathQA~\cite{amini2019mathqa}.

For code generation, we use HumanEval~\cite{chen2021humaneval}, APPS~\cite{hendrycks2021apps}, BigCodeBench~\cite{zhuo2024bigcodebench}, HumanEval+~\cite{liu2023humaneval+}, and MBPP~\cite{austin2021mbpp}.

For medical question answering, we use MedMCQA~\cite{pal2022medmcqa}, MedQA-USMLE~\cite{jin2021usmle}, PubMedQA~\cite{jin2019pubmedqa}, MedQA~\cite{jin2021usmle}, and MMLU~\cite{hendrycks2009mmlu}.

For all benchmarks, prompts are constructed according to their standard evaluation protocols.

\subsection{Training Data}

We use a unified data budget of \textbf{68,000 training samples} for each training stage and domain to ensure consistent settings and comparable training cost.

For the base draft model with shared experts, we use \textbf{ShareGPT}~\cite{ShareGPT}. For mathematical reasoning, we use \textbf{DeepMath-103K}~\cite{he2025deepmath}. For code generation, we use \textbf{Magicoder-Evol-Instruct-110K}~\cite{wei2024magicoder}. For medical question answering, we use the \textbf{Medical} dataset~\cite{medical} and extract English medical QA samples.

\section{Architecture-Level Generalization}
\label{app:llama}

\begin{table}[t]
  \centering
  \caption{Generalization to LLaMA3.1-8B-Math. We report average acceptance length $\tau$ and decoding speedup.}
  \label{tab:llama_generalization}
  \footnotesize
  \setlength{\tabcolsep}{5pt}
  \renewcommand{\arraystretch}{1.05}
  \begin{tabular}{lccccc}
    \toprule
    Method & GSM8K & MathQA & SVAMP & AIME24 & Avg. \\
    \midrule
    LoRA & 4.12 / 2.18$\times$ & 4.75 / 2.07$\times$ & 3.69 / 2.26$\times$ & 3.58 / 2.16$\times$ & 4.04 / 2.17$\times$ \\
    Full-FT & 4.20 / 2.36$\times$ & 4.84 / 2.12$\times$ & 3.75 / 2.30$\times$ & 3.66 / 2.19$\times$ & 4.11 / 2.24$\times$ \\
    EDA & \textbf{4.64 / 2.48$\times$} & \textbf{5.22 / 2.44$\times$} & \textbf{4.12 / 2.52$\times$} & \textbf{3.91 / 2.41$\times$} & \textbf{4.47 / 2.46$\times$} \\
    \bottomrule
  \end{tabular}
\end{table}

To examine whether EDA depends on a specific model family, we evaluate it on LLaMA3.1-8B-Math. This setting differs from the Qwen2.5 family used in our main experiments and provides a test of whether the shared/private draft adaptation principle can transfer across model families.

As shown in Table~\ref{tab:llama_generalization}, EDA consistently outperforms Full-FT and parameter-matched LoRA across all evaluated math benchmarks. These results suggest that EDA is not merely a Qwen-specific adaptation recipe, but can also transfer to another Transformer backbone.

\section{Detailed Analysis of Data Selection}
\label{app:data_selection}

Figure~\ref{fig:selection_metrics} compares our PCA-based representation-shift metric with several training-free baselines. Random sampling uniformly selects adaptation samples. Low-confidence selection ranks samples by the negative average top-1 probability of the target model over answer tokens, favoring samples where the target model is less confident. High-entropy selection ranks samples by the average entropy of the target model's next-token distributions, favoring samples with more uncertain predictions. PCA-reversed selects samples with the smallest representation shifts, serving as the opposite ordering of our method.

As shown in Figure~\ref{fig:selection_metrics}, our method reaches $\tau=4.79$ using only 50\% of the adaptation data, which is close to the full-data performance of 4.83. In contrast, random sampling obtains 4.65 at 50\% and requires 75\% of the data to reach 4.75. Low-confidence and high-entropy selection also underperform our method under the same data budgets, while PCA-reversed performs worst, reaching only 4.50 at 50\%.

These results suggest that uncertainty alone is not the best indicator of sample value for draft adaptation. Low-confidence and high-entropy selection capture local prediction difficulty, whereas our representation-shift metric directly measures deviations from the general reference distribution, better matching the goal of adapting the private expert to target-specific shifts. Moreover, our method is training-free and reuses hidden states collected during self-generation, introducing negligible additional overhead.

\section{Qualitative Analysis}

\begin{figure}[t]
  \centering
  \includegraphics[width=0.6\columnwidth]{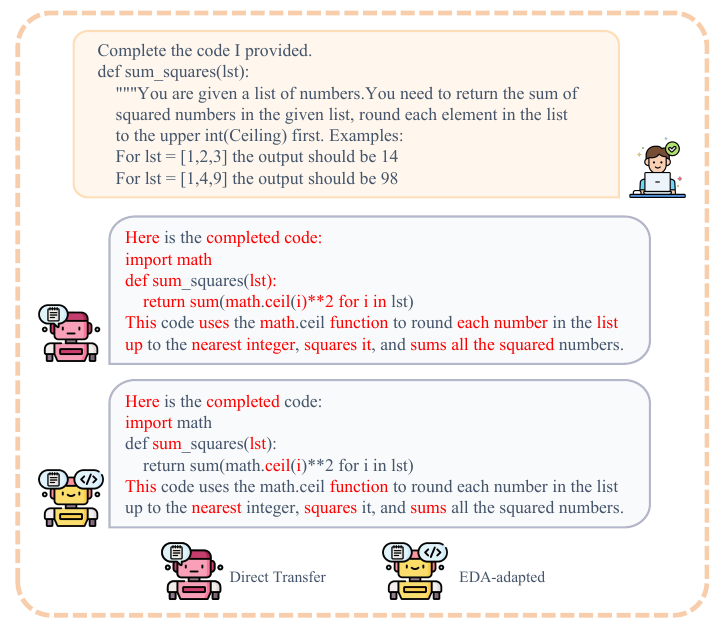}
  \caption{Qualitative example comparing direct transfer and EDA-adapted drafts on Qwen2.5-Coder-7B. Red tokens indicate incorrect predictions, while black tokens indicate correct predictions.}
  \label{fig:chatbot-qualitative}
\end{figure}

To better understand where EDA improves draft-target alignment, we conduct a qualitative case study by comparing direct transfer without adaptation and EDA-adapted drafts.

As shown in Figure~\ref{fig:chatbot-qualitative}, direct transfer often diverges from the target's code-generation trajectory and triggers early rejections. After EDA adaptation, the draft produces more code-consistent tokens, such as \texttt{def} and \texttt{return}, leading to longer consecutive acceptances and more accurate outputs.

We attribute this gain to EDA's efficient realignment: it reuses transferable regularities while lightly adapting target-specific capacity to capture fine-tuning-induced domain conventions. This suggests that EDA mainly corrects preference shifts caused by fine-tuning and formatting conventions, rather than re-learning general language patterns.

\section{Limitations}

Our method is mainly designed for models within the same or closely related model family. Its effectiveness may be limited when applied across substantially different model families or architectures, where the learned representations and optimization behaviors can differ significantly. Extending the proposed method to more diverse model families remains an interesting direction for future work.

\end{document}